\documentclass[11pt]{article}
\usepackage{authblk}
\usepackage{acl2015}
\usepackage{times}
\usepackage{url}
\usepackage{latexsym}
\usepackage{graphicx}
\usepackage{listings}
\usepackage{tabularray}
\usepackage{subfig}

\usepackage{pifont}
\newcommand{\cmark}{\ding{51}}%
\newcommand{\xmark}{\ding{55}}%

\usepackage{amssymb}
\usepackage{amsmath}

\DeclareMathOperator*{\argmax}{arg\,max}

\usepackage[sorting=none]{biblatex}
\title{Tackling VQA with Pretrained Foundation Models without Further Training}

\author[1]{Alvin De Jun Tan*}
\author[2]{Bingquan Shen}
\affil[1]{Nanyang Technological University, Singapore}
\affil[2]{DSO National Laboratories, Singapore}

\date{}

\begin{document}
\maketitle

\begin{abstract}
  Large language models (LLMs) have achieved state-of-the-art results in many natural language processing tasks. They have also demonstrated ability to adapt well to different tasks through zero-shot or few-shot settings. With the capability of these LLMs, researchers have looked into how to adopt them for use with Visual Question Answering (VQA). Many methods require further training to align the image and text embeddings. However, these methods are computationally expensive and requires large scale image-text dataset for training. In this paper, we explore a method of combining pretrained LLMs and other foundation models without further training to solve the VQA problem. The general idea is to use natural language to represent the images such that the LLM can understand the images. We explore different decoding strategies for generating textual representation of the image and evaluate their performance on the VQAv2 dataset. 
\end{abstract}

\section{Introduction}
\def\thefootnote{*}\footnotetext{Work done while author was an intern at DSO National Laboratories.}

In recent years, there has been a remarkable improvement in many natural language processing tasks, driven mainly by the rise of large language models (LLMs). The performance gains offered by these LLMs (e.g. GPT-3 \cite{brown2020language}, Flan-T5 \cite{chung2022scaling}, Llama2 \cite{touvron2023llama}, OPT \cite{zhang2022opt}, etc) can be attributed to scaling up the size of the models and the training data used. These LLMs have also demonstrated ability to perform well in zero-shot and few-shot settings, i.e. the model can quickly adapt to new tasks by providing it a few in-context examples. However, tasks like Visual Question Answering (VQA) requires integrating data from different modalities such as language and images. LLMs alone are not able to solve the VQA task. 

To leverage the performance of pretrained language models for vision-language tasks like VQA, many existing methods require further training to adapt them for visual understanding. A large dataset of image-text pairs are also required in order to align the image and text embeddings. For example, Frozen \cite{tsimpoukelli2021multimodal} and ClipCap \cite{mokady2021clipcap} trains a vision encoder to learn image embeddings that are compatible with a frozen pretrained language model. Flamingo \cite{alayrac2022flamingo} and BLIP2 \cite{li2023blip} uses a specially designed cross-attention module to fuse the respective embeddings from the pretrained vision encoder and pretrained language model. However, the training process required to align the image and text embedding space is very computationally expensive. More recently, there is another line of work that attempts to solve the VQA task by combining together multiple pretrained models without requiring further parameter training or fine-tuning \cite{yang2022empirical, tiong2022plug, guo2022images}.

In this project, we focus on attempting to fuse the vision and language modality by using pretrained language models and vision-language models to tackle the VQA task without any further training. We then study the effect of using different decoding methods during the captioning step to evaluate the performance. The advantage of such a method is that it is flexible and modular, allowing users to easily switch between different models as more powerful ones are released. 

The rest of this paper is organized as follows. In Section 
\ref{sec:related work}, we briefly introduce some related works that solves the VQA task. In Section \ref{sec:method}, we describe our method and the various decoding strategies used during the captioning step. In Section \ref{sec:experiments}, we describe our experimental setup. In Section \ref{sec:results and discussions}, we present our results and some qualitative examples to gain some insights into the performance of the various methos. In Section \ref{sec:limitations and future work}, we highlight some limitations and future work. Lastly, in Section \ref{sec:conclusion}, we conclude with a summary. 

\begin{table*}[ht]
\centering
\caption{Comparison of our results to other methods.}
\label{table:comparison}
\begin{tblr}{
  column{even} = {c},
  column{3} = {c},
  column{5} = {c},
  vline{2-5} = {-}{},
  hline{1-2,11,17} = {-}{},
}
Methods                           & {Extra Multi-Modal\\ Pretraining?} & {Image\\Representation} & {Number \\ of Shots} & {VQAv2 Val\\ Accuracy} \\
Frozen \cite{tsimpoukelli2021multimodal}                           & \cmark                         & Feature Emb.             & 0                    & 29.5                   \\
VL-T5 \cite{cho2021unifying}                             & \cmark                         & Feature Emb.             & 0                    & 13.5                   \\
$FewVLM_{base}$ \cite{jin-etal-2022-good}                  & \cmark                         & Feature Emb.             & 0                    & 43.4                   \\
$FewVLM_{base}$ \cite{jin-etal-2022-good}                 & \cmark                         & Feature Emb.             & 16                   & 48.2                   \\
$FewVLM_{large}$ \cite{jin-etal-2022-good}                 & \cmark                         & Feature Emb.             & 0                    & 47.7                   \\
$FewVLM_{large}$ \cite{jin-etal-2022-good}                & \cmark                         & Feature Emb.             & 16                   & 51.1                   \\
VLKD ViT-B/16 \cite{dai-etal-2022-enabling}                    & \cmark                         & Feature Emb.             & 0                    & 38.6                   \\
VLKD ViT-L/14 \cite{dai-etal-2022-enabling}                    & \cmark                         & Feature Emb.             & 0                    & 42.6                   \\
BLIP-2 ViT-g Flan-T5-XXL \cite{li2023blip}         & \cmark                         & Feature Emb.             & 0                    & 65.2                   \\
PICa \cite{yang2022empirical}                             & \xmark                             & Natural Lang.           & 16                   & 54.3                   \\
PICa-Ensemble \cite{yang2022empirical}                    & \xmark                             & Natural Lang.           & 80                   & 56.1                   \\
Img2LLM \cite{guo2022images}                          & \xmark                             & Natural Lang.           & 0                    & 60.6                   \\
PNP-VQA \cite{tiong2022plug}                          & \xmark                             & Natural Lang.           & 0                    & 63.3                   \\
Ours-\textbf{stochastic}          & \xmark                             & Natural Lang.           & 0                    & 60.23                  \\
Ours-\textbf{\textbf{stochastic}} & \xmark                             & Natural Lang.           & 16                   & 62.39                  
\end{tblr}
\end{table*}

\begin{table*}[ht]
\centering
\caption{Results using the various decoding strategies for captioning.}
\label{table:main results}
\begin{tblr}{
  vline{2-7} = {-}{},
  hline{1-2,6} = {-}{},
}
{Captioning \\ Method}         & n=0   & n=2   & n=4   & n=6   & n=8   & n=16  \\
\textbf{greedy}                & 52.14 & 55.47 & 56.11 & 56.91 & 57.16     & 57.7     \\
\textbf{greedy-tags}           & 51.94 & 54.73 & 55.7  & -     & -     & -     \\
\textbf{stochastic}            & 60.23 & 61.93 & 62.17 & 62.3  & 62.36 & 62.39 \\
\textbf{stochastic-summarized} & 57.47 & 59.22 & -     & -     & -     & -     
\end{tblr}
\end{table*}
 
\section{Related Work}
\label{sec:related work}

\subsection{Visual and Language Pretraining} \label{sec:visual and language pretraining}
There are many approaches proposed recently to align the image and text embeddings. For example, Frozen \cite{tsimpoukelli2021multimodal} uses a pretrained language model and adapts it to align with the vision encoder. Output from the vision encoder is added to the front of the text as the prompt to the frozen language model. Only the vision encoder is updated during training. FewVLM \cite{jin-etal-2022-good} finetunes a sequence-to-sequence transformer model using prefix language modelling and masked language modelling objectives. VLKD \cite{dai-etal-2022-enabling} using knowledge distillation technique to distill multimodal knowledge from CLIP \cite{radford2021learning} to BART \cite{lewis-etal-2020-bart}, a pretrained language model. BLIP2 \cite{li2023blip} uses a specially designed and lightweight Q-former module to connect a frozen vision encoder to a frozen LLM. Different from all the above mentioned work, we study a method that tackles VQA by using pretrained foundation models without further training.

\subsection{Intermediate Representation using Natural Language}
Methods mentioned in Section \ref{sec:visual and language pretraining} uses vectorized representations to represent the image. With the rise of large foundation models and their state-of-the-art performance, there has been more research into methods that no long require computationally expensive training to align the image and text embeddings. To tackle VQA, the general idea is to use natural language as the intermediate representation of the image, through the use of a captioning model to convert the image to text. For example, PICa \cite{yang2022empirical} uses VinVL \cite{zhang2021vinvl} to convert the image to caption and adopts GPT-3 \cite{brown2020language} to answer the question. PNP-VQA \cite{tiong2022plug} identifies regions in the image that are relevant to the question and generates multiple captions based on those regions, then prompt a specialized question-answering model (UnifiedQAv2 \cite{khashabi2022unifiedqa}) to answer the question. In addition to generating question relevant captions, Img2LLM \cite{guo2022images} further generates synthetic question-answer pairs that could be used as in-context examples to prompt a LLM for an answer. 

Such a method is more flexible and modular, enabling the ease of switching between models when newer ones are released. Furthermore, we are able to tap into the implicit knowledge base inherent in the LLM. 

\section{Method} \label{sec:method}

In VQA, we are given open-ended questions about an image. These questions would require an understanding of the image, language and commonsense knowledge to correctly answer it. In this study, we largely follow the method in PICa \cite{yang2022empirical} to tackle the VQA task. To bridge the modality disconnect between the text and image, we resort to natural language as the intermediate representation to interface between the different pretrained models with no further training required. The general idea is to use a captioning model (pretrained vision-language model e.g. BLIP2 \cite{li2023blip}) to generate caption(s) for an image and then prompt a LLM (e.g. Flan-T5 \cite{chung2022scaling}) to give an answer given the caption(s) and the question. In-context examples can also be provided in the prompt to the LLM. This section will introduce in detail the different methods that were tested in our study. 

\subsection{Image Captioning Module}

The image contains a lot of information that would be helpful in answering the question. However, an LLM is unable to directly take in the image as an input. Hence, we use a captioning model (BLIP2 \cite{li2023blip}) that would describe what is happening in the image and generate text that would be used as input to the LLM. We simply provide the image as an input without any text prompt to BLIP2 and BLIP2 is able to generate a caption of the image. We tested different decoding strategies to generate the captions. 

\textbf{Greedy Search}. Greedy search is the simplest decoding strategy, where at each time step $t$, the word $w_t$ with the highest probability is selected as the next word, conditioned on all the previous words $w_{1:t-1}$ and the image $x$: 
\[
w_t = \argmax_w P(w|w_{1:t-1}, x)
\]

\textbf{Greedy Search with Tags}. Similar to PICa \cite{yang2022empirical}, we also enhance the caption generated from greedy search with tags. We prompted the OWL-ViT \cite{owlvit} model with all the 1203 LVIS \cite{gupta2019lvis} category names. We only kept those detected objects with confidence $\ge$ 0.4. The list of category names for each image were appended to the back of the caption, e.g. ``\textless caption\textgreater. \textless tag1\textgreater, \textless tag2\textgreater, \textless tag3\textgreater.". If multiple objects of the same category name was detected, the category name was only appended once.

\textbf{Stochastic Top-$k$ Sampling}. Motivated by \cite{tiong2022plug} and \cite{guo2022images}, given that there may be many things happening and many different ways of describing the same image, we applied stochastic top-$k$ sampling repeatedly. This means that only the $k$ most likely next words are kept and the probability distribution is redistributed among the $k$ next words. We then sample the next word $w_t$ from the new probability distribution at time step $t$, conditioned on all the previous words $w_{1:t-1}$ and the image $x$:
\[
w_t \sim P(w|w_{1:t-1}, x)
\]
This would produce multiple captions for the same image, allowing for a variety and maximal coverage of the image. All the multiple captions were concatenated into a single string, e.g. ``\textless caption1\textgreater. \textless caption2\textgreater. \textless caption3\textgreater. .... \textless caption $N$\textgreater.". In our experiments, we set $k$ = 50 and $N$ = 20.  

\textbf{Summarizing Stochastic Top-$k$ Captions}. After obtaining multiple captions through stochastic top-$k$ sampling, we further investigated a method to summarize all the multiple captions while attempting to keep all the details and descriptions as much as possible. We used Llama-2-13b-chat \cite{touvron2023llama} to summarize the multiple captions generated, using the prompt below:

\begin{lstlisting}[breaklines=true]
<s>[INST] <<SYS>> Below contains some sentences that describes the same image. Write a response that summarizes all the sentences. Keep all details and descriptions as much as possible. <</SYS>>
1. <caption1>.
2. <caption2>.
.
.
.
20. <caption20>.

The above sentences describe the same image. Summarize the sentences above. Keep all the details and descriptions as much as possible. Begin your answer after the word ``answer:" [/INST]
\end{lstlisting}
To force the Llama2 model to give us the answer in a consistent format, we added the last line in our prompt for it to begin the answer after the word ``answer:". The word ``answer:" was removed before using the summarized caption given by Llama2.

\subsection{In-context Examples}

As shown in previous studies \cite{brown2020language, tsimpoukelli2021multimodal}, by providing more in-context examples, the few-shot performance can improve. Following PICa \cite{yang2022empirical}, we select our in-context examples by using the features generated from the CLIP \cite{radford2021learning} model (ViT-B/16 variant). More specifically, given an image $i$ and question $q$ during inference, we forward propagate them through the CLIP model to obtain the image and text embeddings respectively. We compute the cosine similarity of the question with all the other questions in the available in-context examples (from the train set). Similarly, we compute cosine similarity of the image with all the other images in the available in-context examples (from the train set). Then, we average the question similarity with the image similarity to guide the in-context example selection. The top-$n$ image-question pairs with the highest similarity will form the in-context examples for $i$ and $q$.

\begin{table}[ht]
\centering
\caption{Comparison of in-context selection methods. Note that this was done using \textbf{greedy} captioning.}
\label{comparison of in-context selection methods}
\begin{tblr}{
  vline{2-4} = {-}{},
  hline{1,4} = {-}{0.08em},
  hline{2} = {-}{},
}
{In-context Example\\Selection Methods} & n=2   & n=4   & n=6   \\
Question                                & 55.3  & 56.08 & 56.47 \\
Image+Question                          & 55.47 & 56.11 & 56.91 
\end{tblr}
\end{table}

\subsection{Question Answering Module}

We select Flan-T5-XXL \cite{chung2022scaling} as the LLM to prompt for an answer. Flan-T5 is an instruction-finetuned language model that has shown strong few-shot performance compared to even larger pretrained language models. An advantage of Flan-T5 is that it uses relative position encoding \cite{shaw-etal-2018-self}, so theoretically it could generalise to sequences of unseen length. However, we note that as sequence length increases, we are quickly constrained by the memory requirements. The prompt template that was used in our experiment is shown below: 

\begin{lstlisting}
    Context: <caption>
    Question: <question>
    Short Answer: <answer>
    .
    .
    .
    Context: <caption>
    Question: <question>
    Short Answer: 
\end{lstlisting}
The answer was produced in an open-ended generation manner, in contrast to some other methods that tackle VQA as a classification task and assumes access to answer candidates \cite{changpinyo-etal-2022-may, banerjee-etal-2021-weaqa}, which is not realistic in the real world. Note that we converted all the predicted answers generated by Flan-T5 to lowercase.

\section{Experiments} \label{sec:experiments}

\subsection{Dataset and Evaluation}

In this study, we evaluated on the VQAv2 \cite{goyal2017making} validation set. VQAv2 consists of question-answer pairs based on the COCO image dataset, consisting of 214,354 and 443,757 questions in the validation and train set respectively. The questions in VQAv2 were designed to be relevant to the image content. We evaluated our answers based on open-ended generation and report the soft-accuracy, given by:
\[ Acc(\text{ans}) = min\{\frac{\text{\# of humans that said ans}}{3}, 1\} \]
To be consistent with ``human accuracies", machine accuracies are averaged over all 10 choose 9 sets of human annotators.

\begin{figure*}
    \centering
    \includegraphics[width=\textwidth]{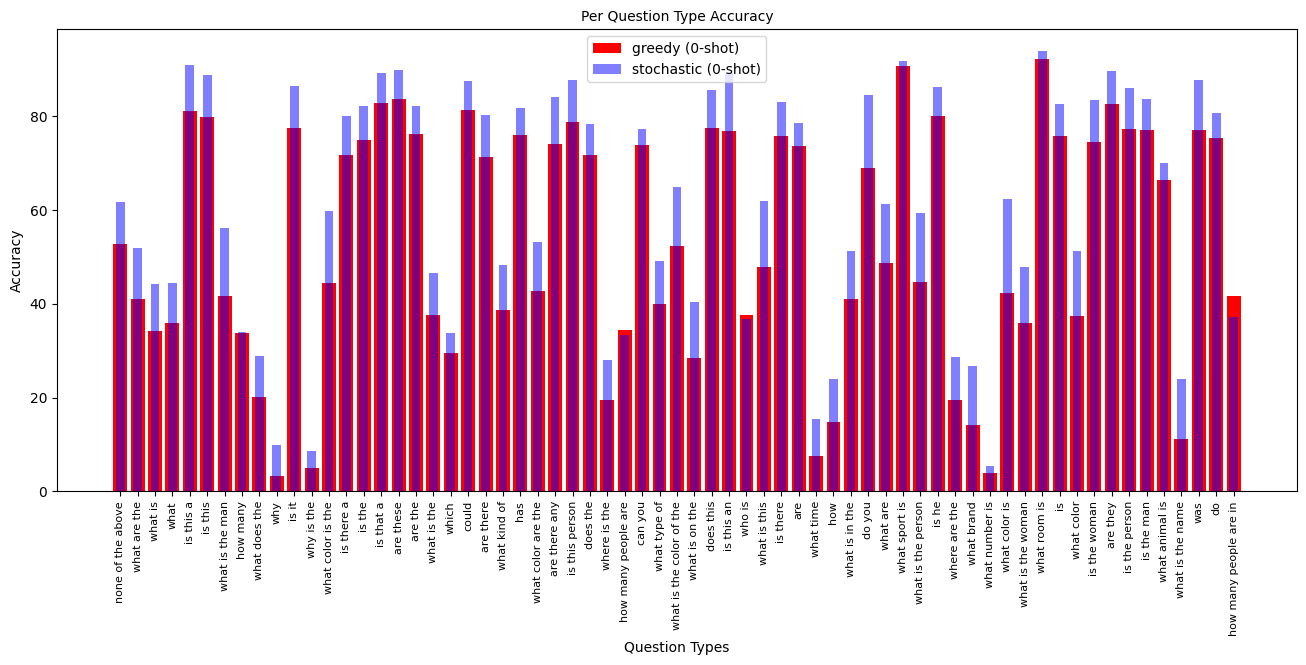}
    \caption{Per question accuracy for \textbf{greedy (0-shot)} vs \textbf{stochastic (0-shot).}}
    \label{fig:per question accuracy greedy vs stochastic}
\end{figure*}

\subsection{Setup}

The models used in our implementation are all available from HuggingFace. For BLIP2, we used the model name ``Salesforce/blip2-opt-2.7b-coco"; for Flan-T5-XXL we used ``google/flan-t5-xxl"; for Llama2 we used ``meta-llama/Llama-2-13b-chat-hf". 

We compared different variants of the captioning method for the image during inference time in our experiments:
\begin{itemize}
    \item \textbf{greedy}. Caption was generated using the greedy search strategy. 
    \item \textbf{greedy-tags}. Tags, generated by prompting OWL-ViT with LVIS cateogry names, were appended to the back of the \textbf{greedy} caption. 
    \item \textbf{stochastic}. We concatenated all the 20 captions that were independently sampled using stochastic top-$k$ method into a single string. 
    \item \textbf{stochastic-summarized}. Using the 20 captions generated using stochastic top-$k$ sampling, we prompted Llama-2-13b-chat to give a detailed summary. 
\end{itemize}
Note that all the captions used as the ``Context" for the in-context examples were generated using greedy search to reduce computational complexity and memory constraints. The decoding strategy used for Flan-T5 was greedy with a maximum new tokens of 5 for \textbf{greedy} and \textbf{greedy-tags}, while we used beam search with a width of 5 and length penalty of -1 for \textbf{stochastic} and \textbf{stochastic-summarized}. 

\section{Results and Discussions} \label{sec:results and discussions}

\begin{figure*}
    \centering
    \includegraphics[width=\textwidth]{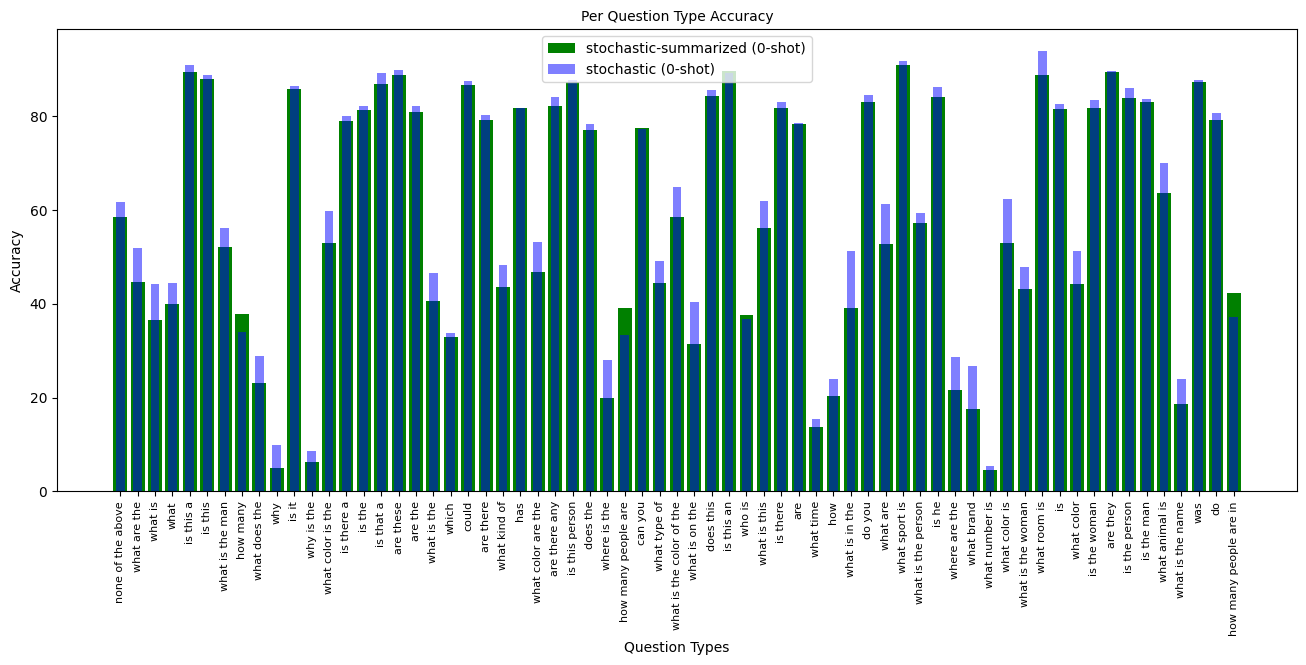}
    \caption{Per question accuracy for \textbf{stochastic-summarized (0-shot)} vs \textbf{stochastic (0-shot).}}
    \label{fig:per question accuracy stochastic-summarized vs stochastic}
\end{figure*}

\subsection{Main Results}
Table \ref{table:main results} summarizes the main results of our study. We roughly summarize some general trends in this section. 

\textbf{More In-context Examples Improves Performance}. We observe that throughout all the different captioning methods, more in-context examples do help achieve higher performance. However, the improvement diminishes as the number of in-context examples increases. This is most evidently seen in the case of \textbf{stochastic} method where increasing the number of shots from 8 to 16 only leads to a marginal 0.03 increase (62.36 vs 62.39). 

\textbf{LVIS Tags Leads to Lower Performance}. Based on Table \ref{table:main results}, comparing \textbf{greedy-tags} and \textbf{greedy}, we observe that there is a slight drop in the accuracy when tags were used. In contrast to the results from PICa where tags from the Microsoft Azure tagging API increases their performance, by appending the LVIS object category names into the caption made the accuracy lower in our case. We suspect that appending the LVIS category names to the captions may be adding more noise rather than guiding the language model towards the correct answer.

\textbf{Stochastic Top-$k$ Leads to Better Performance}. We observe that by independently sampling 20 captions using the stochastic top-$k$ method and concatenating them into one string, we achieve much higher performance than just using greedy caption, suggesting that this method could enrich the description and understanding of the image, thereby providing more details that may be relevant to answering the question. For zero-shot, the \textbf{stochastic} method outperforms the \textbf{greedy} method by 8.09\%. More analysis of this will be done in Section \ref{sec:stochastic over greedy}.

\textbf{Summarizing Stochastic Top-$k$ Leads to Lower Performance}. We observe a lower accuracy by using the summarized version compared to simply concatenating the 20 captions sampled independently using stochastic top-$k$ into a single string. This could be due to the loss of information from paraphrasing as the Llama2 model was performing the summary. More analysis of this will be done in Section \ref{sec:stochastic-summarized over stochastic}.

\subsection{Comparison to Other Methods}
We compared our results to other methods in Table \ref{table:comparison}, roughly grouping them into two categories: (i) Methods that require multi-modal pretraining objective to align the text and image embeddings, such as Frozen \cite{tsimpoukelli2021multimodal}, VL-T5 \cite{cho2021unifying}, FewVLM \cite{jin-etal-2022-good}, VLKD \cite{dai-etal-2022-enabling} and BLIP2 \cite{li2023blip}. These methods are computationally expensive and require a large-scale visual-language datasets to train. Some of these methods, e.g. FewVLM, also reported their accuracies using few-shot learning. (ii) Methods that use frozen pretrained models without further training, such as PICa \cite{yang2022empirical}, Img2LLM \cite{guo2022images}, PNP-VQA \cite{tiong2022plug}. Our method belongs to this category. 

Using the \textbf{stochastic} captioning method, we managed to outperform PICa by a large margin, in both zero-shot and few-shot settings. Even with 0-shot \textbf{stochastic} method, we outperform the 80-shot PICa-Ensemble method. We note that this is likely due to the fact that we used better captioning model, i.e. BLIP2 vs VinVL in PICa, and better language model, i.e. Flan-T5-XXL vs GPT-3 in PICa. This also illustrates the advantage of using such a modular system, where we can easily switch between models as more powerful ones are released to improve the performance. Our 0-shot \textbf{stochastic} method also outperforms all of the methods identified in Table \ref{table:comparison} that requires extra multi-modal pretraining, except for BLIP-2 ViT-g Flan-T5-XXL. 
 
\subsection{Comparison of In-Context Selection Method}
We report the results on selecting the in-context examples through using only question similarity compared to image-question averaged similarity. Note that we conducted the experiments using the \textbf{greedy} captioning method. From Table \ref{comparison of in-context selection methods}, we observe only a marginal improvement of selecting the in-context examples from using the averaged image-question similarity compared to only using question similarity.

\subsection{Improvement of stochastic over greedy}
\label{sec:stochastic over greedy}
To further analyse how the \textbf{stochastic} top-$k$ captioning method improves the performance, we  plot the per question type accuracy for \textbf{greedy (0-shot)} vs \textbf{stochastic (0-shot)} in Figure \ref{fig:per question accuracy greedy vs stochastic}. In general, the \textbf{stochastic} method outperforms the \textbf{greedy} method across all question types, except for ``how many people are" and ``how many people are in" (an explanation will be discussed in \ref{sec:stochastic-summarized over stochastic}). We provide some qualitative examples in Figure \ref{fig:qualitative examples of stochastic vs greedy}. By independently sampling multiple captions of the same image, we are able to get more details about the image, each time potentially focusing on different aspects in the image. For example in Figure \ref{fig:qualitative examples of stochastic vs greedy}(a), the \textbf{greedy} method fails to capture the buildings/houses in the background, and hence no relevant details are available to help Flan-T5 in answering the question of ``what is behind the trees?". On the other hand, the \textbf{stochastic} top-$k$ method is able to capture the buildings/houses in some of the captions, thus providing more details that could guide Flan-T5 in generating the correct answer. 

\subsection{Lower Performance of stochastic-summarized over stochastic}
\label{sec:stochastic-summarized over stochastic}
Again, we plot the per question type accuracy for \textbf{stochastic-summarized (0-shot)} vs \textbf{stochastic (0-shot)} in Figure \ref{fig:per question accuracy stochastic-summarized vs stochastic}. Across all the question types, we observe that in general, summarizing the 20 captions generated by \textbf{stochastic} top-$k$ method leads to lower accuracy (except for questions that begins with ``how many", ``how many people are", ``how many people are in" etc). Figure \ref{fig:qualitative examples of stochastic vs stochastic-summarized worst} shows some qualitative examples of how the \textbf{stochastic-summarized} caption performs worst than \textbf{stochastic}. For example in Figure \ref{fig:qualitative examples of stochastic vs stochastic-summarized worst}(a), we can observe some form of ``hallucination" by Llama2. Although there was no mention of ``african savannah" in the 20 independently sampled captions, the summarized version generated by Llama2 mentioned that ``The image captures the beauty and diversity of the African savannah". This could have potentially led to Flan-T5 generating the answer ``african savannah" to the question ``where are the giraffes?". In another example in Figure \ref{fig:qualitative examples of stochastic vs stochastic-summarized worst}(b), Llama2 failed to capture in the summary that the white bird in the image was a ``seagull", which was mentioned in one of the 20 independently sampled captions, leading to Flan-T5 giving an incorrect answer. 
% In Figure \ref{fig:qualitative examples of stochastic vs stochastic-summarized worst}(c) and \ref{fig:qualitative examples of stochastic vs stochastic-summarized worst}(d), we observe that the summarized version may have paraphrased some of the descriptions 

Next, we take a closer look into the qualitative examples with questions that begin with ``how many" in Figure \ref{fig:qualitative examples of stochastic vs stochastic-summarized how many question types}. For example in Figure \ref{fig:qualitative examples of stochastic vs stochastic-summarized how many question types}(c), in each of the captions sampled from \textbf{stochastic} top-$k$, it is quite clear that there is 1 young/little boy in the image. However, the answer predicted by Flan-T5 to the question ``how many children are shown?" was 4, suggesting that concatenating all the captions into one string makes it difficult  for the language model to understand how many of the object in question is being described. This could add more noise, hence confusing Flan-T5 when trying to understand the context. On the other hand, the summarized version makes the description of the image more coherent, guiding Flan-T5 towards the correct prediction of 1. This would also possibly explain the lower performance of \textbf{stochastic} compared to \textbf{greedy} in the ``how many" question types as previously pointed out in Section \ref{sec:stochastic over greedy}.

\section{Limitations and Future Work} \label{sec:limitations and future work}

\subsection{Limitation of VQA Soft-Accuracy Metric}
We would like to point out a limitation of the soft-accuracy metric used to evaluate VQA performance. Due to the way the metric was formulated, after processing\footnote{The processing steps can be found in: https://visualqa.org/evaluation.html}, the generated answer have to match at least one of the ground truth answers by the human annotators, otherwise it will be scored 0. This leads to many of the answers generated to be marked as incorrect, although they may be logically correct. We provide some qualitative example of such cases in Figure \ref{fig:qualitative examples of metric limitations}. For example, in Figure \ref{fig:qualitative examples of metric limitations}(b), both of the answers ``to keep warm" and ``it is cold outside" are logically correct answers to ``why are the people wearing hats?". However, they are both scored 0, due to the fact that they do not match at least one of the ground truth answers. This could partially explain the low accuracy for question types beginning with ``why" as shown in Figures \ref{fig:per question accuracy greedy vs stochastic} and \ref{fig:per question accuracy stochastic-summarized vs stochastic}. Due to the subjective nature, there may be many different ways of answering the question. Certain answers that are logically correct will be incorrectly penalised under such a metric.

\subsection{Limitation of Captioning as Image Representation}
From Figures \ref{fig:per question accuracy greedy vs stochastic} and \ref{fig:per question accuracy stochastic-summarized vs stochastic}, we also noticed poor performance in question types ``what number is" and ``what time". In general, we noticed that the captions generated by BLIP2 do not detect text/numbers and time on the clock well. In most of the cases, the time on the clock are not described in the captions generated, resulting in the low accuracy for such question types. This could possibly be explained by the fact that BLIP2 was not trained for detecting time on a clock or detecting text/numbers in images (although BLIP2 do exhibit OCR capabilities \cite{liu2023hidden}). Another explanation could be the fact that the clock or text/numbers in the images were not the main focus in most cases, hence the captions generated will not include the time or text/numbers.

\subsection{Future Work}
More research have to be done to find ways to boost the performance of the ``why", ``what number is", ``what time", etc type of questions. As a first step, we could try to adapt the method described in \cite{tiong2022plug} and \cite{guo2022images} to extract question relevant patches and generate captions based on those patches using BLIP2. This could help boost the accuracy as the captions are now based on image regions that may be relevant to the question, as compared to generating captions based on the whole image with no particular aim. Next, we could also independently sample more captions using the \textbf{stochastic} top-$k$ method, and use a larger Llama2 model (e.g. Llama2-70b-chat) or ChatGPT to summarize the captions. Lastly, we expect that a end-to-end pretraining (to align the image and text modality) and then finetuning on VQAv2 directly will help achieve higher performance.

\section{Conclusion} \label{sec:conclusion}
In conclusion, in this work, we attempted to build a system similar to PICa using BLIP2 as the captioner and Flan-T5 as the LLM to solve the VQA task without further training. We explored various decoding methods for the captioning step, such as the greedy method, repeatedly sampling using stochastic top-$k$, and then using Llama2 to summarize the multiple captions generated from stochastic top-$k$. We found that concatenating the multiple captions sampled using stochastic top-$k$ into a single string performs the best, and even outperforms PICa. We provided some qualitative analysis into the performance of the various captioning methods and highlighted some limitations of such a modular system. 

\printbibliography

\begin{figure*}
    \centering
    \includegraphics[width=\textwidth]{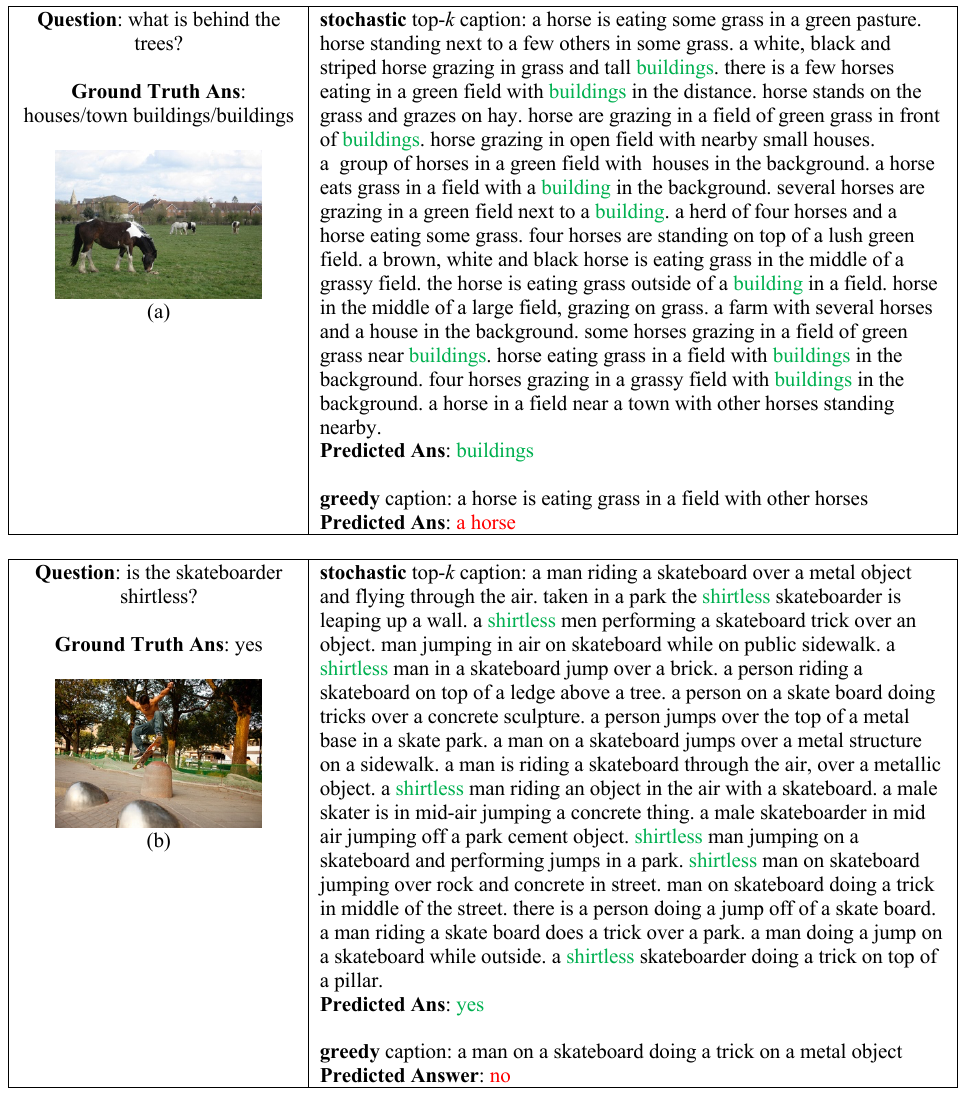}
    \caption{Qualitative examples of how \textbf{stochastic} top-$k$ captions perform better than \textbf{greedy} captions. The answer cues and correct answers are indicated by green color, while red color indicates wrong answers. (a): the \textbf{greedy} caption fails to capture the buildings in the background, whereas the \textbf{stochastic} method is able to capture the buildings in some of the captions. (b): \textbf{greedy} caption fails to capture that the man is shirtless, whereas the \textbf{stochastic} method is able to capture that the man is shirtless in some of the captions.}
    \label{fig:qualitative examples of stochastic vs greedy}
\end{figure*}
\begin{figure*} \ContinuedFloat
    \centering
    \includegraphics[width=\textwidth]{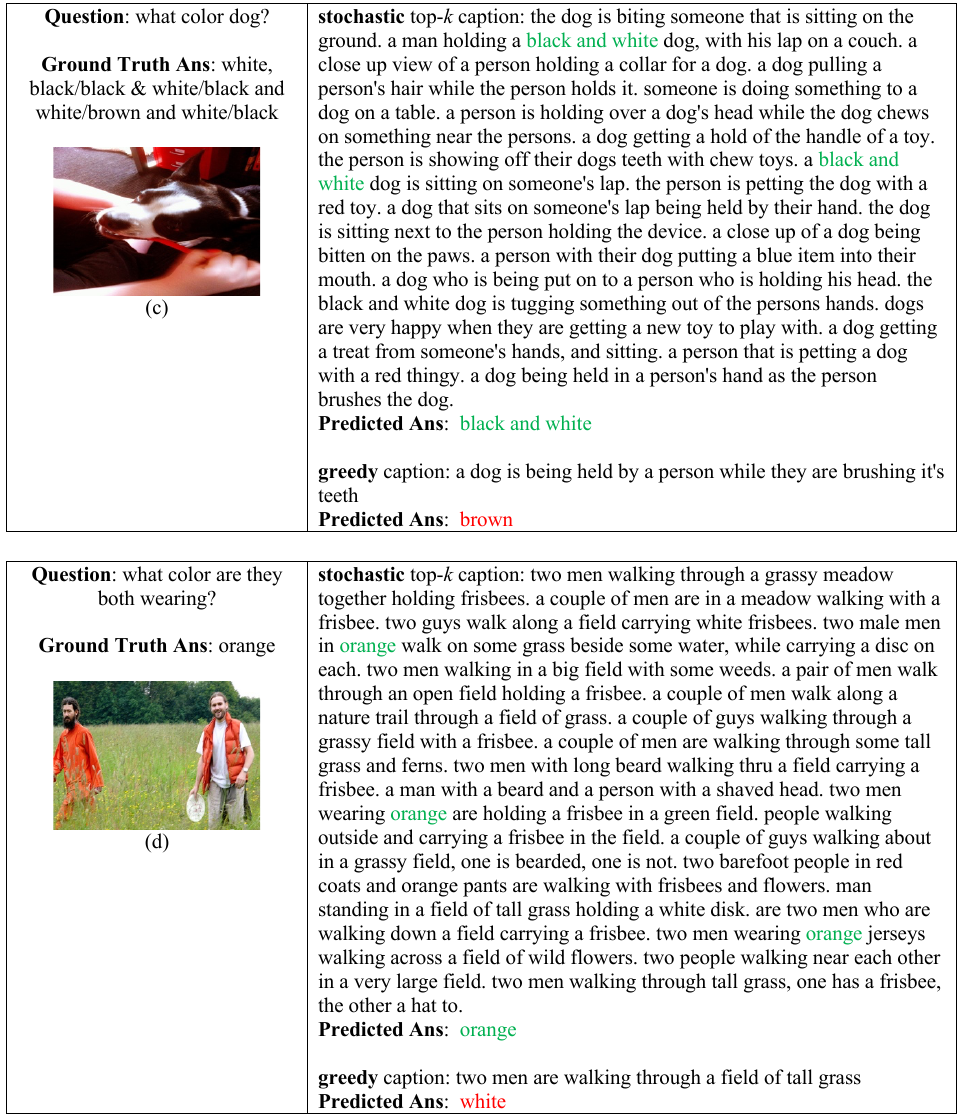}
    \caption{Qualitative examples of how \textbf{stochastic} top-$k$ captions perform better than \textbf{greedy} captions. The answer cues and correct answers are indicated by green color, while red color indicates wrong answers. (c): the \textbf{greedy} caption does not describe the color of the dog, whereas the \textbf{stochastic} method is able to capture the color of the dog in some of the captions. (d): the \textbf{greedy} caption does not capture the color of the clothing, whereas the \textbf{stochastic} method captures the color of the clothing in both of the man in some of the captions.}
    \label{fig:qualitative examples of stochastic vs greedy}
\end{figure*}

\begin{figure*}
    \centering
    \includegraphics[width=\textwidth]{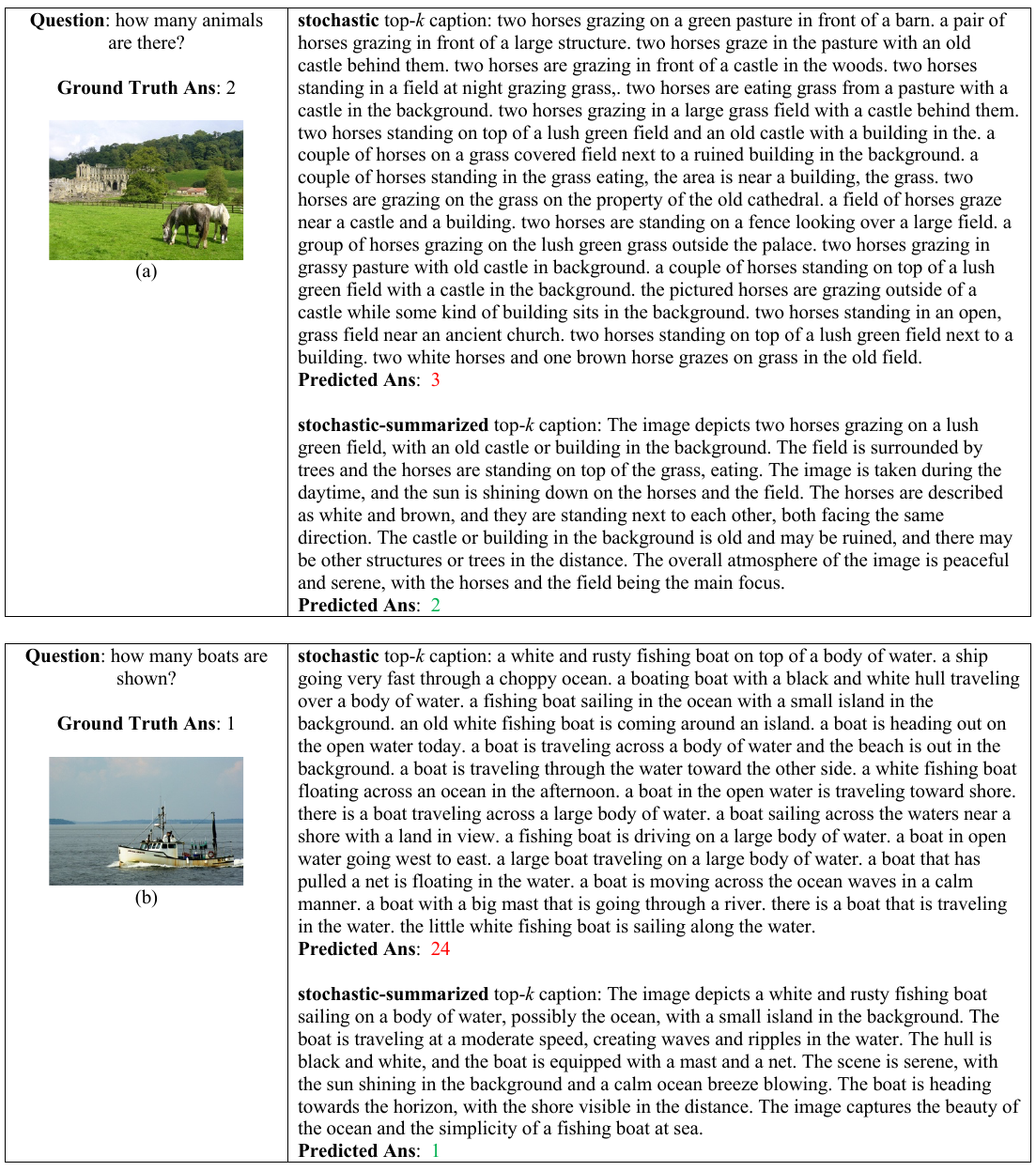}
    \caption{Qualitative examples of how \textbf{stochastic-summarized} captions perform better than \textbf{stochastic} captions for ``how many" question types. Green color indicates correct answer predicted, while red color indicates wrong answer predicted. Simply concatenating the 20 captions from \textbf{stochastic} top-$k$ into one string seems to confuse Flan-T5 as to how many objects in question are in the image.}
    \label{fig:qualitative examples of stochastic vs stochastic-summarized how many question types}
\end{figure*}
\begin{figure*} \ContinuedFloat
    \centering
    \includegraphics[width=\textwidth]{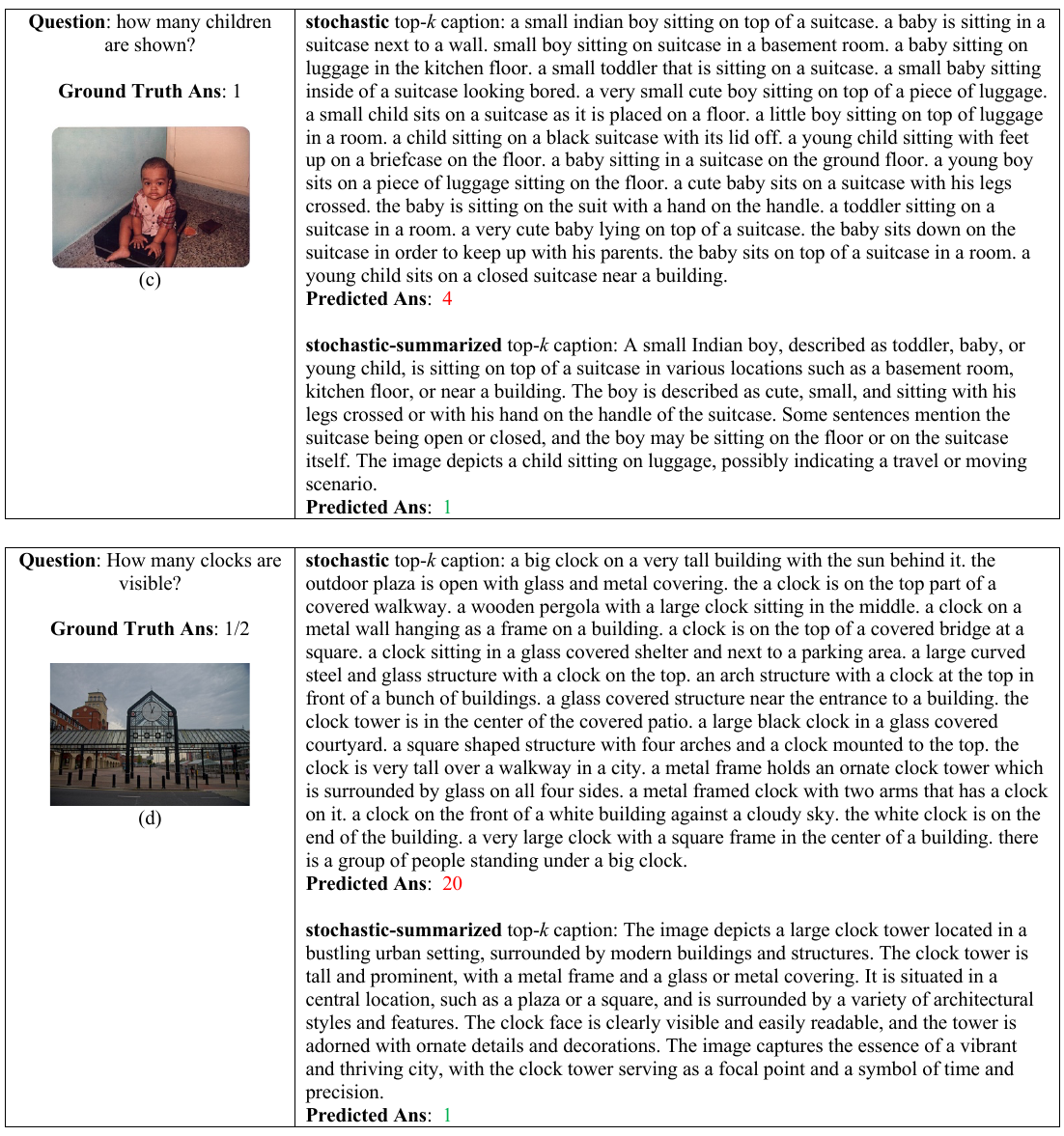}
    \caption{Qualitative examples of how \textbf{stochastic-summarized} captions perform better than \textbf{stochastic} captions for ``how many" question types. Green color indicates correct answer predicted, while red color indicates wrong answer predicted. Simply concatenating the 20 captions from \textbf{stochastic} top-$k$ into one string seems to confuse Flan-T5 as to how many objects in question are in the image.}
    \label{fig:qualitative examples of stochastic vs stochastic-summarized how many question types}
\end{figure*}

\begin{figure*}
    \centering
    \includegraphics[width=\textwidth]{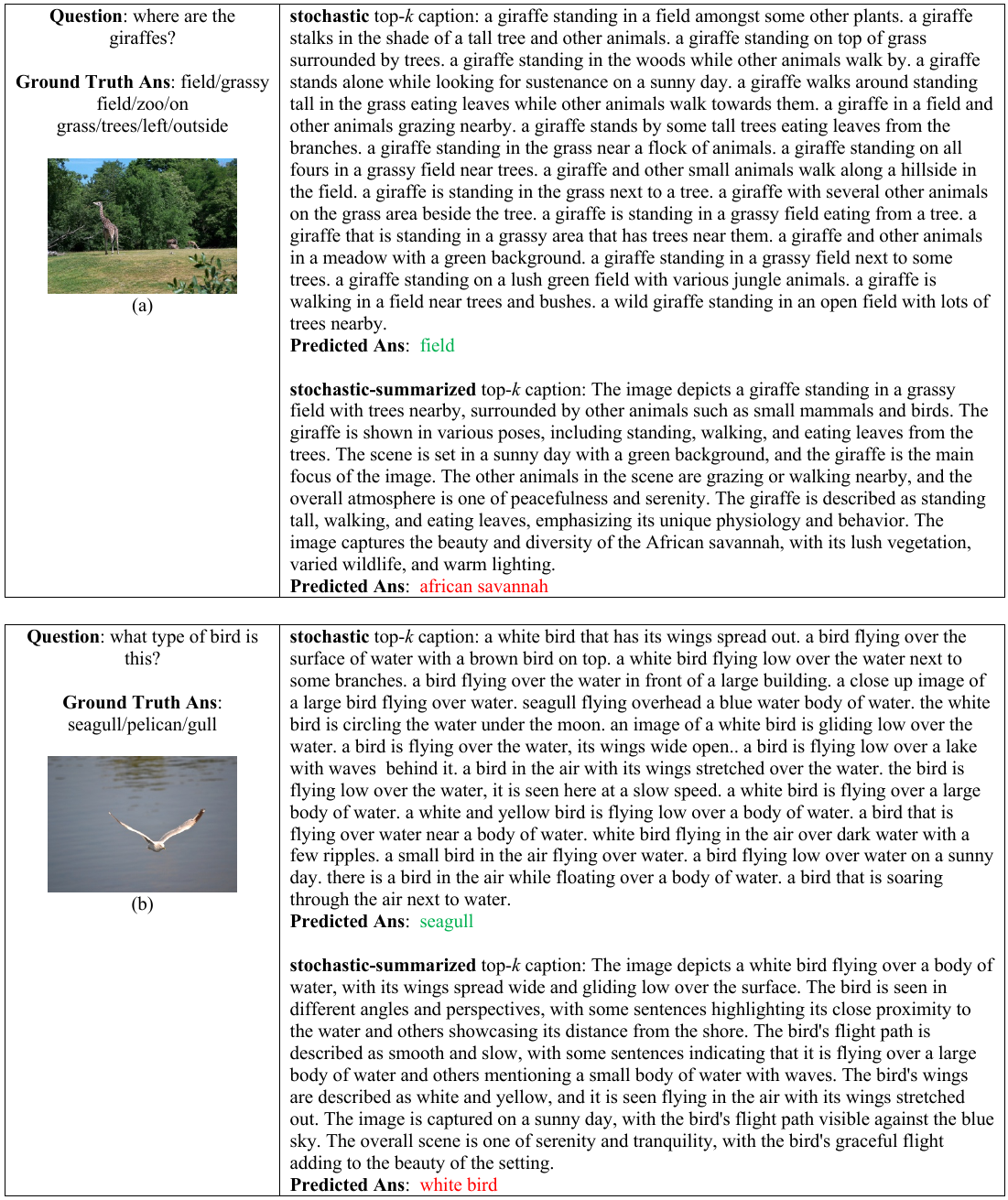}
    \caption{Qualitative examples of how \textbf{stochastic-summarized} captions perform worst than \textbf{stochastic} captions. Green color indicates correct answer predicted, while red color indicates wrong answer predicted. (a): Llama2 ``hallucinating" in its summary and adding the description of the location ``African savannah" when it did not appear in the captions generated by \textbf{stochastic} top-$k$. (b): Llama2 fails to capture the ``seagull" in its summary when it appeared in one of the captions generated by \textbf{stochastic} top-$k$.}
    \label{fig:qualitative examples of stochastic vs stochastic-summarized worst}
\end{figure*}
\begin{figure*} \ContinuedFloat
    \centering
    \includegraphics[width=\textwidth]{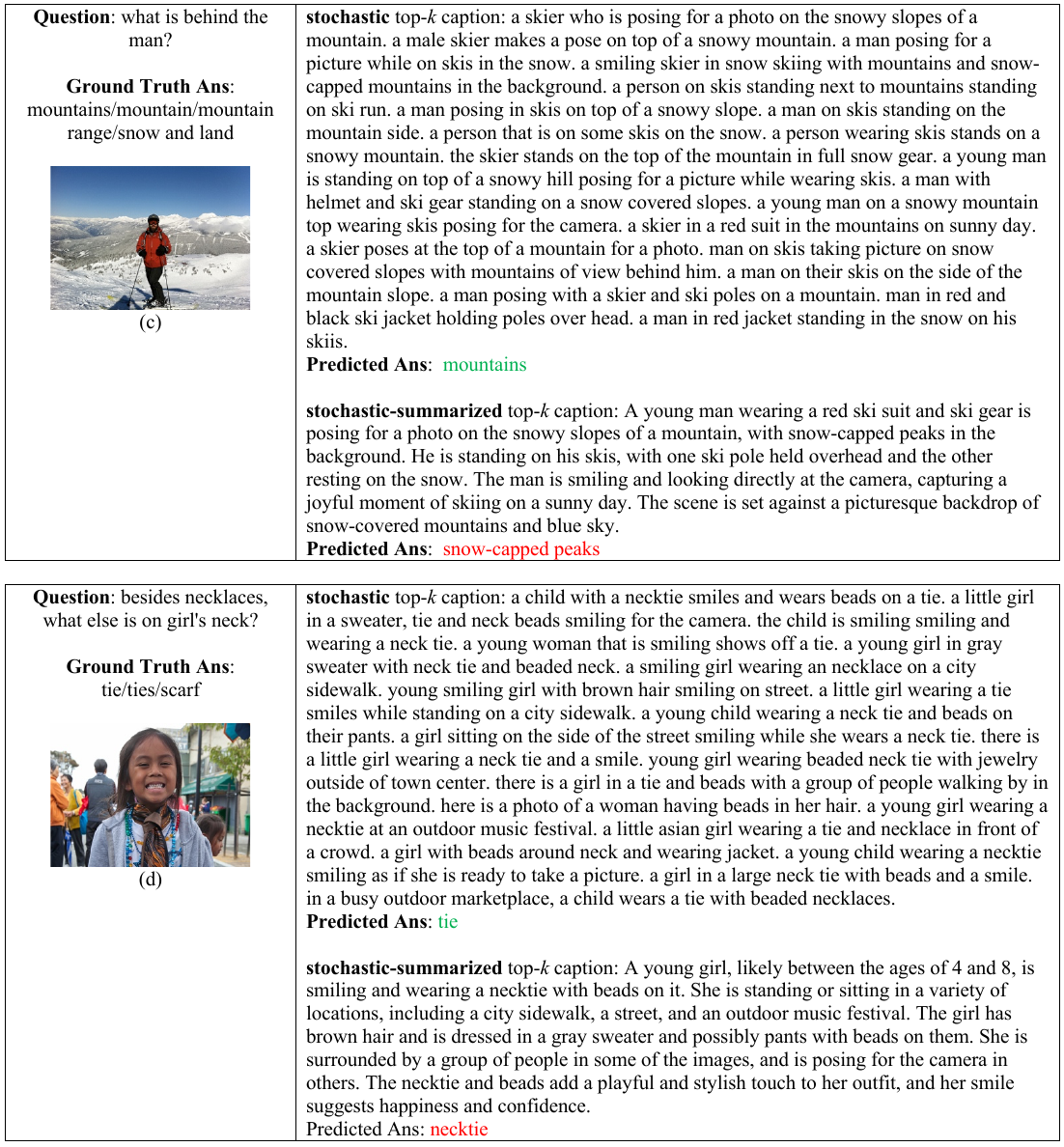}
    \caption{Qualitative examples of how \textbf{stochastic-summarized} captions perform worst than \textbf{stochastic} captions. Green color indicates correct answer predicted, while red color indicates wrong answer predicted. (c): Llama2 paraphrases the description of the snowy mountains in the background as ``snow-capped peaks", leading to a ``wrong" answer generated (although logically it should be correct, see a discussion of the limitation of the evaluation metric in Section \ref{sec:limitations and future work}). (d): Llama2 uses the word ``necktie" to summarize that the young girl is wearing a tie, leading to the ``wrong" answer of necktie (again, see a discussion of the limitation of the evaluation metric in Section \ref{sec:limitations and future work}).}
    \label{fig:qualitative examples of stochastic vs stochastic-summarized worst}
\end{figure*}

\begin{figure*}
    \centering
    \includegraphics[width=\textwidth]{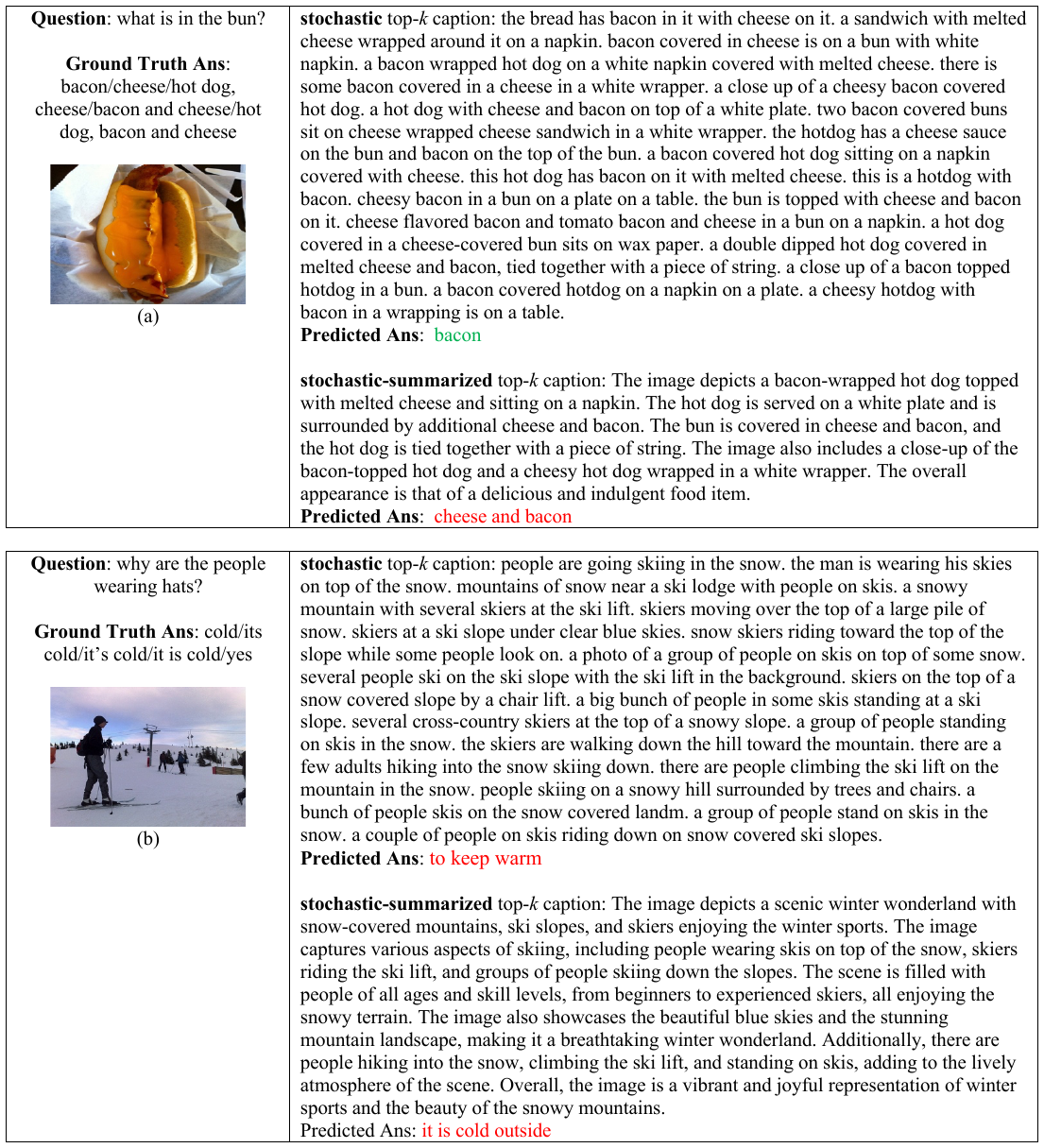}
    \caption{Qualitative examples to illustrate the limitations of soft-accuracy to evaluate VQA performance. Green color indicates correct answer predicted, while red color indicates wrong answer predicted.}
    \label{fig:qualitative examples of metric limitations}
\end{figure*}
\begin{figure*} \ContinuedFloat
    \centering
    \includegraphics[width=\textwidth]{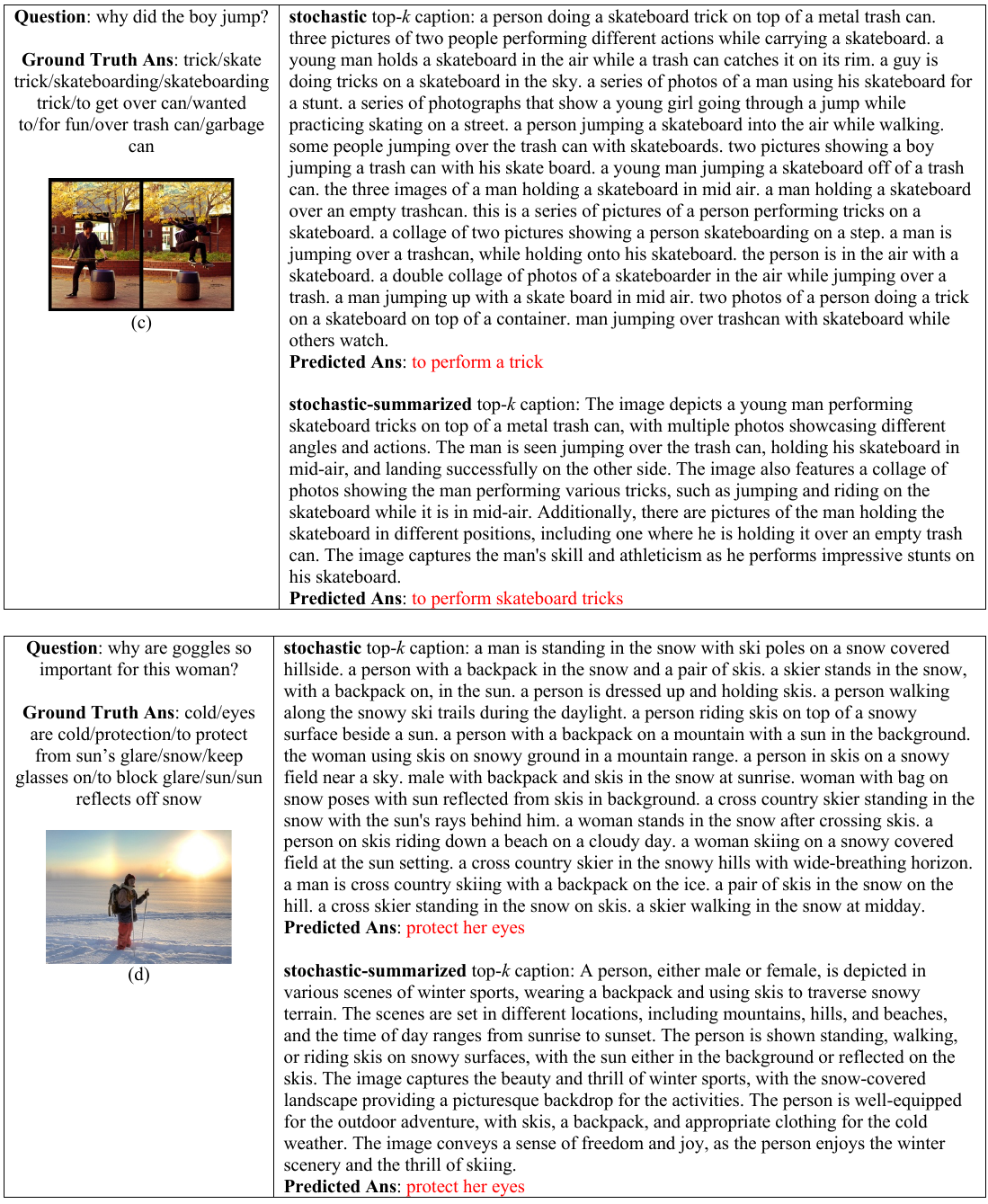}
    \caption{Qualitative examples to illustrate the limitations of soft-accuracy to evaluate VQA performance. Green color indicates correct answer predicted, while red color indicates wrong answer predicted.}
    \label{fig:qualitative examples of metric limitations}
\end{figure*}

\end{document}